\documentclass[10pt, conference, compsocconf]{IEEEtran}
\ifCLASSINFOpdf
\else
\fi

\usepackage{booktabs} 
\usepackage{graphicx}
\usepackage{capt-of}
\usepackage{epstopdf}
\usepackage[linesnumbered,ruled]{algorithm2e}
\usepackage{algorithmic}
\usepackage{colortbl}
\usepackage{xcolor}
\usepackage{subcaption}
\usepackage{amssymb}
\usepackage{amsmath}
\usepackage{float}
\usepackage[export]{adjustbox}
\usepackage{authblk}

\let\OLDthebibliography\thebibliography
\renewcommand\thebibliography[1]{
  \OLDthebibliography{#1}
  \setlength{\parskip}{0pt}
  \setlength{\itemsep}{0pt plus 0.3ex}
}

\begin{document}
\title{Scalable Bottom-up Subspace Clustering using FP-Trees for High Dimensional Data}

\author[*1]{Minh Tuan Doan}
\author[1]{Jianzhong Qi}
\author[2]{Sutharshan Rajasegarar}
\author[*1]{Christopher Leckie}
\affil[*]{\fontsize{10}{12}\selectfont Data61, CSIRO, Australia}
\affil[1]{\fontsize{10}{12}\selectfont School of Computing and Information Systems, The University of Melbourne, Australia}
\affil[2]{\fontsize{10}{12}\selectfont School of Information Technology, Deakin University, Australia}
\affil[ ]{\fontsize{10}{12}\selectfont Email: \{mdoan@student., jianzhong.qi, caleckie\}@unimelb.edu.au\\sutharshan.rajasegarar@deakin.edu.au }

\maketitle

\begin{abstract}
Subspace clustering aims to find groups of similar objects (clusters) that exist in lower dimensional subspaces from a high dimensional dataset. It has a wide range of applications, such as analysing high dimensional sensor data or DNA sequences. However, existing algorithms have limitations in finding clusters in non-disjoint subspaces and scaling to large data, which impinge their applicability in areas such as bioinformatics and the Internet of Things. We aim to address such limitations by proposing a subspace clustering algorithm using a bottom-up strategy. Our algorithm first searches for base clusters in low dimensional subspaces. It then forms clusters in higher-dimensional subspaces using these base clusters, which we formulate as a frequent pattern mining problem. This formulation enables efficient search for clusters in higher-dimensional subspaces, which is done using FP-trees. The proposed algorithm is evaluated against traditional bottom-up clustering algorithms and state-of-the-art subspace clustering algorithms. The experimental results show that the proposed algorithm produces clusters with high accuracy, and scales well to large volumes of data. We also demonstrate the algorithm's performance using real-life data, including ten genomic datasets and a car parking occupancy dataset.
\end{abstract}

%
%


\section{Introduction}
Subspace clustering aims to find groups of similar objects, or clusters, that exist in lower dimensional subspaces from a high dimensional dataset. This has a wide range of applications, including the rapidly growing fields of the Internet of Things (IoT) \cite{jin2014information} and bioinformatics~\cite{chen2017subspace}. Applications such as these generate large volumes of high dimensional data, which bring new challenges to the subspace clustering problem. In this paper we propose a novel approach to subspace clustering that addresses two key challenges in these applications: scalability to large datasets and non-disjoint subspaces.

The first challenge lies in handling large inputs. This is essential for many applications nowadays since the captured data can grow to million of records in a short period of time. It has been shown \cite{muller2009evaluating,vidal2011subspace} that many existing algorithms have high computational costs and take considerable time to cluster relatively small inputs, e.g., STATPC~\cite{muller2009evaluating} needs more than 13 hours to cluster 7,500 records of 16 dimensions.
Table \ref{example_intro} illustrates how our algorithm can scale to inputs with large volumes of data, in comparison to state-of-the-art subspace clustering algorithms SWCC~\cite{chen2017subspace}, SSC~\cite{elhamifar2013sparse}, and LRR~\cite{liu2011latent}. The running time of our algorithm over 100,000 data points is half that required by SWCC (which is a highly efficient co-clustering algorithm, but cannot find clusters in non-disjoint subspaces). The state-of-the-art subspace clustering algorithms SSC and LRR also suffer as the number of data points increases. SSC triggers memory errors when the numbers of data points reaches 15,000, while LRR cannot terminate in 12 hours for just 5,000 points.

\begin{table}[h!]
\centering
\begin{tabular}{lllllll}
         & 5,000  & 10,000 & 15,000  & 20,000 & 50,000 & 100,000 \\ \hline
Ours & \textbf{6.7}   & \textbf{13.2}  & \textbf{20.7}   & \textbf{28.7}  & \textbf{127.9} & \textbf{184.5}  \\
SWCC     & 9.8  & 19.9 & 37.8  & 93.94 & 198.96 & 374.48 \\
SSC      & 226.1 & 416.9 & 1506.4 & -     & -     & -      \\
LRR      & -      & -      & -       & -      & -      & -       \\ \hline
\end{tabular}
\caption{Clustering time (in seconds) on 10-dimensional datasets. The volume ranges from 5,000 to 100,000 points.}
\label{example_intro}
\end{table}

\begin{figure}[h!]
	\vspace{-1em}
	\centering
	\includegraphics[width=0.49\textwidth]{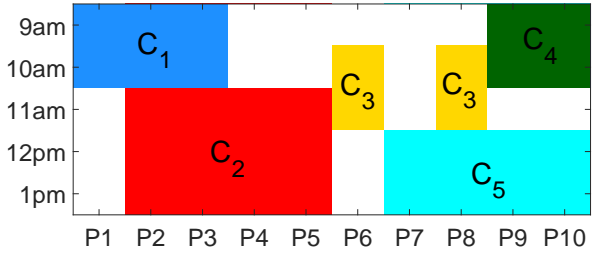}
	\caption{An illustration of clusters in non-disjoint subspaces for car parking occupancy data. Clusters are highlighted to show simultaneous groupings of points and dimensions.}
	\label{heatmap_parking}
\end{figure}

The second challenge involves finding clusters in non-disjoint subspaces \cite{wang2016noisy}. Many recent algorithms \cite{elhamifar2013sparse,liu2011latent} assume that clusters are located in disjoint subspaces, which do not have any intersection except for the origin. This is a strong assumption that can be unrealistic, because real-life data may be correlated in different overlapping subsets of dimensions, also known as the property of \emph{local feature relevance} \cite{kriegel2009clustering}. For example, with gene expression data, a particular gene can be involved in multiple genetic pathways, which can result in different symptoms among different sets of patients \cite{diaz2006gene}. Hence, a gene can belong to different clusters that have dimensions in common while differing in other dimensions \cite{hanisch2002co}. Figure \ref{heatmap_parking} presents another example of clusters in non-disjoint subspaces that are observed in data collected from IoT applications. The heatmap visualizes the subspace clustering results of a car parking occupancy dataset at 10 locations from 9am to 1pm, where each column represents a car parking bay, and each row represents an hour of the day. 
It can be observed that clusters $C_1$ and $C_2$ are in non-disjoint subspaces since they share the dimensions of parking bays \verb!P2! and \verb!P3! in common. In the case of $C_1$, this can be interpreted as the utilisation of these two parking bays following some pattern that is also observed at \verb!P1! between 9am-10am. On the other hand, cluster $C_2$ shows that \verb!P2! and \verb!P3! follow a different pattern between 11am-1pm, and share that pattern with \verb!P4! and \verb!P5!. Further analysis of the data can suggest that $\{P2,P3,P1\}$ are busy parking bays during morning peaks, whereas $\{P2,P3,P4,P5\}$ have higher occupancy levels during lunch time.

To address these challenges, we propose a novel algorithm that can find clusters in non-disjoint subspaces and scale well with large inputs. The algorithm follows a bottom-up strategy and comprises two phases. First, it searches for potential clusters in low dimensional subspaces, which we call \emph{base clusters}. We start with base clusters instead of dense units in separate dimensions, which are used in existing bottom-up clustering algorithms \cite{kriegel2009clustering}. This allows our algorithm to preserve the covariance of data between different dimensions, which is also a critical factor when clustering high dimensional data, as we further elaborate in Section 4.1. In addition, this approach makes our algorithm more stable and tolerant to variations in parameters settings.

\begin{figure}[h!]
	\vspace{-1em}
	\centering
	\includegraphics[width=0.45\textwidth,height=3cm]{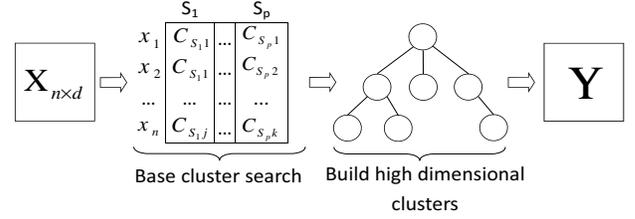}
	\caption{Framework of the proposed algorithm}
	\label{algo_diagram}
	\vspace{-1em}
\end{figure}

In the second phase, base clusters that share similar sets of data points are aggregated together to form clusters in higher dimensional subspaces. This process of aggregation is non-trivial. One of the main challenges lies in keeping the number of aggregated clusters tractable. This not only directly affects the computational costs of the algorithm, but also ensures that the final result is presented in an appropriate number of meaningful clusters.
Many existing algorithms~\cite{Agrawal:1998:ASC:276304.276314,parsons2004subspace} depend on combinatorial search to combine low dimensional clusters (dense units). If there are on average $m$ dense units in each dimension, the first level of aggregation of CLIQUE~\cite{Agrawal:1998:ASC:276304.276314} (to combine one-dimensional dense units into two-dimensional clusters) would need to check $|m|^d$ pairwise possible aggregations, where $d$ is the number of dimensions. Further aggregation would need to be applied sequentially for each subsequent higher dimension. We alleviate this heavy computation by transforming the aggregation problem into a frequent pattern mining problem \cite{han2000mining} to achieve efficient and robust aggregation of base clusters. This approach also allows us to avoid the construction of a similarity matrix, which has quadratic complexity with respect to the input volume. Therefore, we reduce both time and space complexity and enable the algorithm to work with very large inputs. During this process, a base cluster may be aggregated into more than one cluster in different higher dimensional subspaces that have overlapping dimensions, which enables us to find non-disjoint subspace clusters. The general steps of our algorithms are summarized in Figure \ref{algo_diagram} and are detailed in Section 4.

We make the following contributions:
\begin{itemize}
	\item We propose a novel subspace clustering algorithm that can find clusters in non-disjoint subspaces and handle very large inputs. The novelty of our approach is reflected in both phases of the algorithm. First, we search for base clusters in low dimensional subspaces to preserve the covariance of data between different dimensions. Second, we transform the process of sequential aggregation of low dimensional clusters to a problem of frequent pattern mining to construct high dimensional clusters.
	\item We demonstrate that the proposed algorithm outperforms traditional subspace clustering algorithms using bottom-up strategies, as well as state-of-the-art algorithms with other clustering strategies, in terms of accuracy and scalability on large volumes of data.
	\item We conduct a range of experiments to demonstrate the effectiveness of our algorithm in different practical applications. Specifically, we present how the algorithm can be applied to (1) real-life sensor data from the City of Melbourne, Australia \cite{Melbourne_car_parking:url}, and (2) 10 different gene expression datasets \cite{diaz2006gene}, and produce comparable or better results than state-of-the-art algorithms.
\end{itemize}

\section{Related work}
Subspace clustering is an active research field that aims to partition high dimensional datasets into groups of objects that are similar in subspaces of the data space. The attributes of high dimensional data lead to multiple challenges for subspace clustering. A major challenge is referred to as \emph{local feature relevance} \cite{kriegel2009clustering}, which states that clusters only exist in subspaces (or subsets of dimensions) rather than the full dimensional space. In addition, the subspaces where a cluster exists vary for different subsets of data points. This phenomenon makes traditional similarity measures, such as Euclidean distance, Manhattan distance, and cosine similarity ineffective. The reason is that these measures use all dimensions, both relevant and irrelevant, when computing similarity. Moreover, since subspaces vary for different (and unknown) subsets of points, common dimensionality reduction techniques, such as PCA \cite{wold1987principal}, MDS~\cite{kruskal1978multidimensional}, and feature selection methods \cite{guyon2003introduction} that apply global changes to the data, are not effective.

\textbf{Subspace clustering methods.}
Subspace clustering methods can be categorised into five groups: iterative methods, algebraic methods, statistical methods, matrix factorisation-based methods, and spectral clustering based methods. We briefly describe each group with representative algorithms. A detailed survey of these algorithms is in \cite{vidal2011subspace}.

Iterative methods suchs as K-subspaces~\cite{wang2009k} iteratively alternate between assigning points to the subspaces and updating subspaces to refine the clusters. K-subspaces is simple, fast, and is guaranteed to converge. However, it needs to know the number of clusters as well as the dimensions of each cluster beforehand. The algorithm is also sensitive to outliers and only converges to a local optimum.

Statistical methods, such as MPPCA \cite{tipping1999mixtures}, assume that the data in each subspace follow a known distribution, such as a Gaussian distribution. The clustering process alternates between clustering the data and adjusting the subspaces by maximizing the expectation of the principle components of all subspaces. These algorithms need to know the number of clusters as well as the number of dimensions of each subspace. Moreover, their accuracy heavily depends on the initialization of the clusters and subspaces.

GPCA \cite{vidal2005generalized} is a representative algorithm of the algebraic methods. It considers the full data space as the union of $s$ underlying subspaces, and hence represents the input data as a polynomial $P(x)$ of degree $s$:
$P(x)=\prod_{i=1}^{s}b_i^\top x=(b_1^\top x)...(b_s^\top x)=0$
where $b_i$ and $b_i^\top x$ are the normal vector and the equation of subspace $S_i$ respectively. The subspaces are then identified by grouping the normal vectors $\vec{n}_i$ of all the points, which are the derivatives at the values $x_i$. GPCA needs to know the number of dimensions of each subspace, and is sensitive to noise and outliers. Besides, GPCA has high computational complexity and does not scale well to the number of subspaces or their dimensionalities.

Matrix factorization based algorithms use low-rank factorization to construct the similarity matrix over the data points. Specifically, given the input $N \in \mathbb{R}^{n\times d}$ containing $n$ points in $d$-dimensions, matrix factorization based algorithms~\cite{boult1991factorization}, \cite{costeira1998multibody} find the SVD \cite{wold1987principal} of the input to subsequently construct the similarity matrix $Z$, where $Z_{ij}=0$ if points $i$ and $j$ belong to different subspaces.
The final clusters are obtained by thresholding the entries of $Z$. These methods assume the subspaces to be independent and noise free.


SSC \cite{elhamifar2013sparse} and LRR \cite{liu2011latent} are two state-of-the-art algorithms that use spectral clustering techniques. They initially express each data point $x_i \in N$ as a linear combination of the remaining data $x_i=\Sigma_{i\neq j}^n z_{ij}x_j$, and use the coefficients $z_{ij}$ to construct the similarity matrix $Z\in \mathbb{R}^{n\times n}$. The algorithms then optimize $Z$ to make $z_{ij}=0$ for all points $x_i,x_j$ that do not belong to the same subspace. SSC uses $L_1$-norm regularization \cite{liu2010efficient} to enforce $Z$ to be sparse, while LRR enforces the matrix to be low-rank by using nuclear norm regularization \cite{jaggi2010simple}. Both algorithms assume the underlying subspaces to be disjoint. In addition, both have high computational complexity, which grows rapidly with the number of input records.

\textbf{Bottom-up subspace clustering algorithms.} From an algorithmic point of view, clustering algorithms can be classified into bottom-up algorithms and top-down algorithms \cite{kriegel2009clustering}. As our algorithm follows a bottom-up strategy, we briefly discuss the relevant algorithms of this class to highlight our contributions.

The bottom-up strategy involves searching for dense units in individual dimensions, and subsequently aggregating these dense units to form clusters in higher dimensional subspaces. The difference among bottom-up algorithms lies in the definition of dense units and the method of aggregating lower dimensional clusters. For example, CLIQUE \cite{Agrawal:1998:ASC:276304.276314} divides individual dimensions into fixed size cells, and defines dense units as cells containing more than a predefined number of points. It then aggregates adjacent dense units to construct higher dimensional clusters. CLIQUE heavily depends on setting appropriate values of the cell size and density threshold. This can be challenging because the value ranges differ in different dimensions and there might not be a single set of parameters that suit all dimensions. In addition, searching for dense units in separate dimensions omits the covariance between dimensions, which can lead to either missing clusters or redundant combinations of dense units. We discuss this phenomenon in more detail in Section 4.1. SUBCLU \cite{parsons2004subspace} does not rely on fixed cells. Instead, it uses DBSCAN \cite{kriegel2009clustering} to search for dense units in each dimension, and iteratively constructs higher dimensional subspaces. The algorithm invokes a call of DBSCAN for each candidate subspace, which can lead to a high running time. We propose to perform clustering only at the beginning of the algorithm while still guaranteeing that the aggregation of these $2$-dimensional clusters form valid high dimensional clusters, which achieves a much lower computational cost.

\textbf{Co-clustering.} Another relevant topic is co-clustering (a.k.a bi-clustering or pattern-based clustering) \cite{aggarwal2013data}. Co-clustering can be considered as a more general class of clustering high dimensional data by simultaneously clustering rows (points) and columns (dimensions). The main point that differentiates co-clustering from subspace clustering lies in the approach to the problem, and the homogeneous methodology to find clusters in both axis-parallel and arbitrarily oriented subspaces~\cite{kriegel2009clustering}.
In this paper, we also compare the performance of our algorithm on gene expression data with a range of co-clustering algorithms, including SWCC \cite{chen2017subspace}, BBAC-S \cite{banerjee2007generalized}, ITCC \cite{dhillon2003information}, FFCFW \cite{tjhi2006flexible}, and HICC \cite{cheng2016hicc}.

\section{Problem statement}
We first present the notation used in this paper.

\begin{itemize}
	\item $S_i^{(k)}$ is a subspace of $k$ dimensions, which is represented as a set of its component dimensions: $S_i^{(k)}=\{d_{i1},...,d_{ij},...,d_{ik}\}$, $d_{ij}$ represents the $j^{th}$ dimension.
	\item $X_j$ or $\{x_j\}$ is a set of points; $x_j$ denotes a point: $x_j=\{x_{ji}\}_{i=1}^{k}$, $x_{ji}$ is the coordinate in the $i^{th}$ dimension.
	\item $C_{S_i}^{X_j}$ is a cluster formed by points $X_j$ in subspace $S_i$.
\end{itemize}

\noindent Let $X=\{x_{i}\in\mathbb{R}^d:i=1..n\}$ be a set of $n$ points in a $d$-dimensional space, and $X_j$ be a subset of $X$. The set of all subspace clusters is denoted as $Y=\{C_{S_i}^{X_j},\; i:1..s,\; j:1..c\}$. Here, $s$ denotes the number of subspaces containing clusters, and $c$ denotes the number of all clusters. More than one cluster can exist in a subspace, i.e., $c\geq s$. \textit{Our subspace clustering algorithm finds all clusters by identifying their corresponding subspaces and point sets.}

We take a bottom-up approach to find the clusters in subspaces starting from finding base clusters in low dimensional subspaces. The algorithm to find the base clusters is orthogonal to our study. We use k-means in the experiments for simplicity, although any low dimensional clustering algorithms may be used.
Once the base clusters are found, our algorithm aggregates them to form clusters in higher-dimensional subspaces. We follow a probabilistic approach together with the downward closure property of density to guarantee the validity of the formation of clusters in higher dimensional subspaces. This is formulated as Lemma~1.

\textbf{Lemma 1: }Given two points $x_1$ and $x_2$ in subspace $S_i$, the probability that $x_1$ and $x_2$ belong to the same cluster in subspace $S_i$ is proportional to the cardinality $|\{S_{i'}\}|$ ($S_{i'} \subset S_i$) in which $x_1$ and $x_2$ belong to the same cluster.

\textbf{Proof: }Let $C_{S_i}$ denote the event where two points $x_1$ and $x_2$ belong to the same cluster in subspace $S_i$. Assume that we already perform clustering in lower dimensional subspaces and find that these two points belong to the same cluster in a set of $p$ subspaces $\mathcal{S}=\{S_{i1}, ..., S_{ij}, ..., S_{ip}\}$ ($S_{ij} \subset S_i$). Given this knowledge, the probability that $x_1$ and $x_2$ belong to the same cluster in $S_i$ is:
$$P_1=P(C_{S_i}|\;C_{S_{i1}},...,C_{S_{ip}})=\frac{P(C_{S_i},C_{S_{i1}},...,C_{S_{ip}})}{P(C_{S_{i1}},...,C_{S_{ip}})}$$

We show that the probability $P_1$ increases as new evidence of the cluster formation of $x_1$ and $x_2$ is found in other subspaces of $S_i$. Specifically, let these two points also belong to a cluster in a certain subspace $S_{im} \subset S_i$ ($S_{im} \not\subset \mathcal{S}$, i.e., $S_{im}$ is indeed a newly discovered subspace in which $x_1$ and $x_2$ belong to the same cluster). The probability of them belonging to the same cluster in $S_i$ becomes:

\vspace{-1em}
{\smaller
\begin{align}
P_2=P(C_{S_i}|\:C_{S_{i1}},...,C_{S_{ip}},C_{S_{im}})=\frac{P(C_{S_i},C_{S_{i1}},...,C_{S_{ip}},C_{S_{im}})}{P(C_{S_{i1}},...,C_{S_{ip}},C_{S_{im}})}\nonumber
\end{align}
}%
\vspace{-1em}

By applying the chain rule, we can show that $P_2>P_1$:
\begin{small}
\begin{align}
\frac{P_2}{P_1}=\frac{P(C_{S_i},C_{S_{i1}},...,C_{S_{ip}},C_{S_{im}})}{P(C_{S_{i1}},...,C_{S_{ip}},C_{S_{im}})} \times \frac{P(C_{S_{i1}},...,C_{S_{ip}})}{P(C_{S_i},C_{S_{i1}},...,C_{S_{ip}})}\nonumber \label{ratio_1}
\end{align}
\end{small}

According to the downward closure property of density, if $x_1$ and $x_2$ are near in $S_{i}$, they are also near in all subspaces of $S_{i}$, including $S_{im}$. Hence, $P(C_{S_{im}}|\;C_{S_{i}})=1$, or $P(C_{S_{im}},C_{S_{i}})=P(C_{S_{i}})$. Therefore, $P(C_{S_i},C_{S_{i1}},...,C_{S_{ip}},C_{S_{im}})=P(C_{S_i},C_{S_{i1}},...,C_{S_{ip}})$. The previous equation can then be rewritten as:
\vspace{-0.3em}
\begin{small}
\begin{align}
\frac{P_2}{P_1}=\frac{P(C_{S_{i1}},...,C_{S_{ip}})}{P(C_{S_{i1}},...,C_{S_{ip}},C_{S_{im}})}=\frac{\sum_{C_{S_{im}}}P(C_{S_{i1}},...,C_{S_{ip}},C_{S_{im}})}{P(C_{S_{i1}},...,C_{S_{ip}},C_{S_{im}})} \nonumber
\end{align}
\end{small}
\vspace{-0.5em}

By marginalising the numerator over $C_{S_{im}}$, we can deduce that $\frac{P2}{P1}\geq 1$. We therefore show that additional evidence of $x_1$ and $x_2$ belonging to the same cluster in another subspace $S_{mi}\subset S_i$ increases the probability that these two points belong to the same cluster in $S_i$. Thus, Lemma 1 is proved.
\parbox[][0.1em]{\textwidth}{}

The intuition of Lemma 1 is that the formation of clusters in lower dimensional subspaces can be used as evidence to reinforce and increase the posterior probability of the formation of a cluster for the same set of points in the higher dimensional super subspaces. Therefore, \emph{we say that there is a high probability that a set of points form a cluster in a high dimensional subspace if they form clusters in a sufficiently large number of its subspaces.} 

\section{Proposed method}
We propose a two-phase subspace clustering algorithm as summarised in Algorithm 1.

\begin{algorithm}[h]
	\small
	\algsetup{linenosize=tiny}
	
    \SetKwInOut{Input}{Input}
    \SetKwInOut{Output}{Output}

    \Input{$X \in \mathbb{R}^{n \times d}$\\
    		$num\_of\_subspaces$, $min\_cluster\_size$
    		}
    \Output{$FP=\{F_i:i=1..q\}$: set of frequent patterns that represent the clusters}
    
	\tcp{Phase 1: search for base clusters in lower dimensional subspaces}
    $Z=0_{n,num\_of\_subspaces}$\\
    \For{i :=1 to $num\_of\_subspaces$}{
    	$S_{sample}\leftarrow sample(S)$ \tcp{{\smaller sample a rand subspace}}
		$Z_{*,i}\leftarrow$ cluster($X_{S_{sample}}$)
	}

	\tcp{Phase 2: Extract clusters from FP-Tree}
	$min\_sup\leftarrow\frac{min\_cluster\_size}{n}$\\
	$T\leftarrow$ build\_fp\_tree($Z$,$min\_sup$)\\
	\tcp{Analyse freq at each level to prune T}
	$T\leftarrow$ prune\_tree($T$)	\\
	
	$\{FP_i,Z_i\}\leftarrow$ extract\_maximal\_frequent\_sets($T$)

	\Return{$\{FP_i,Z_i\}$}

    \caption{Bottom-up clustering using FP-Tree}
\end{algorithm}
\vspace{-0.5em}

\subsection{Phase 1: Base Cluster Search}
\begin{figure*}[t]
	\centering
	\begin{subfigure}[b]{.24\textwidth}
		\centering
		\includegraphics[width=\textwidth]{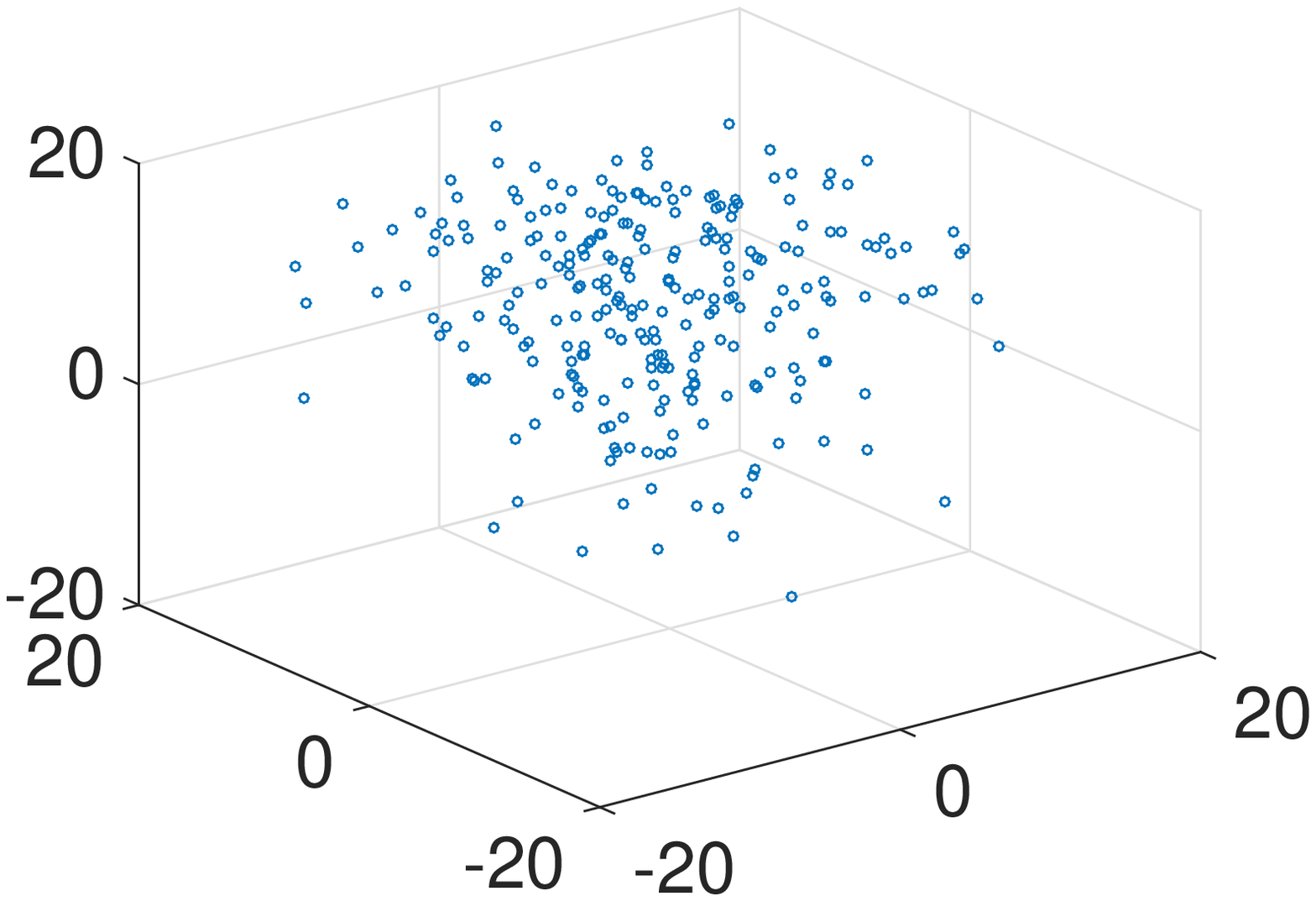}
		\caption{}
		\label{phase1_arg_1}
	\end{subfigure}
	\begin{subfigure}[b]{.24\textwidth}
		\centering
		\includegraphics[width=\textwidth]{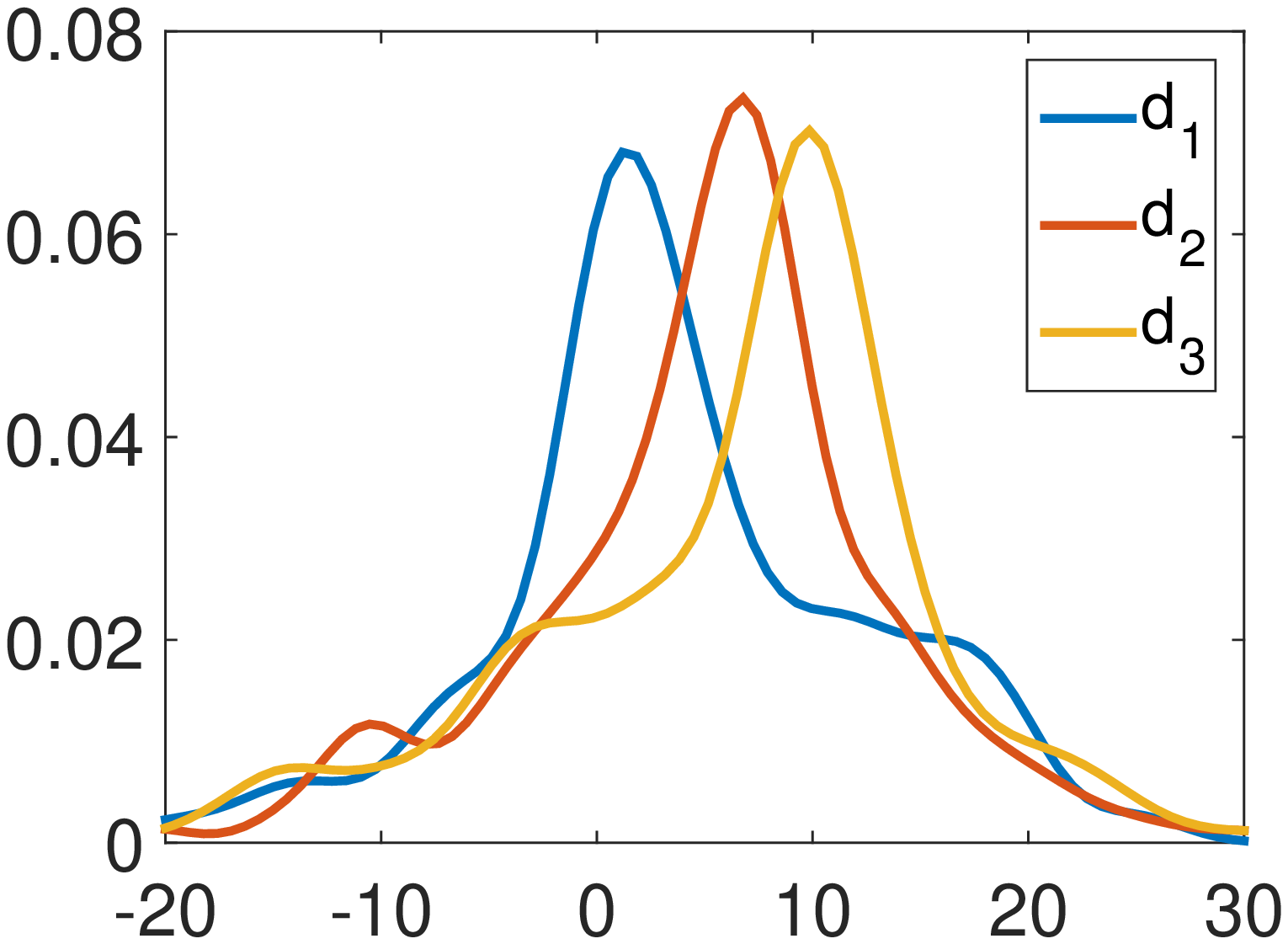}
		\caption{}
		\label{kdensity_1}
	\end{subfigure}
	\begin{subfigure}[b]{.24\textwidth}
		\centering
		\includegraphics[width=\textwidth]{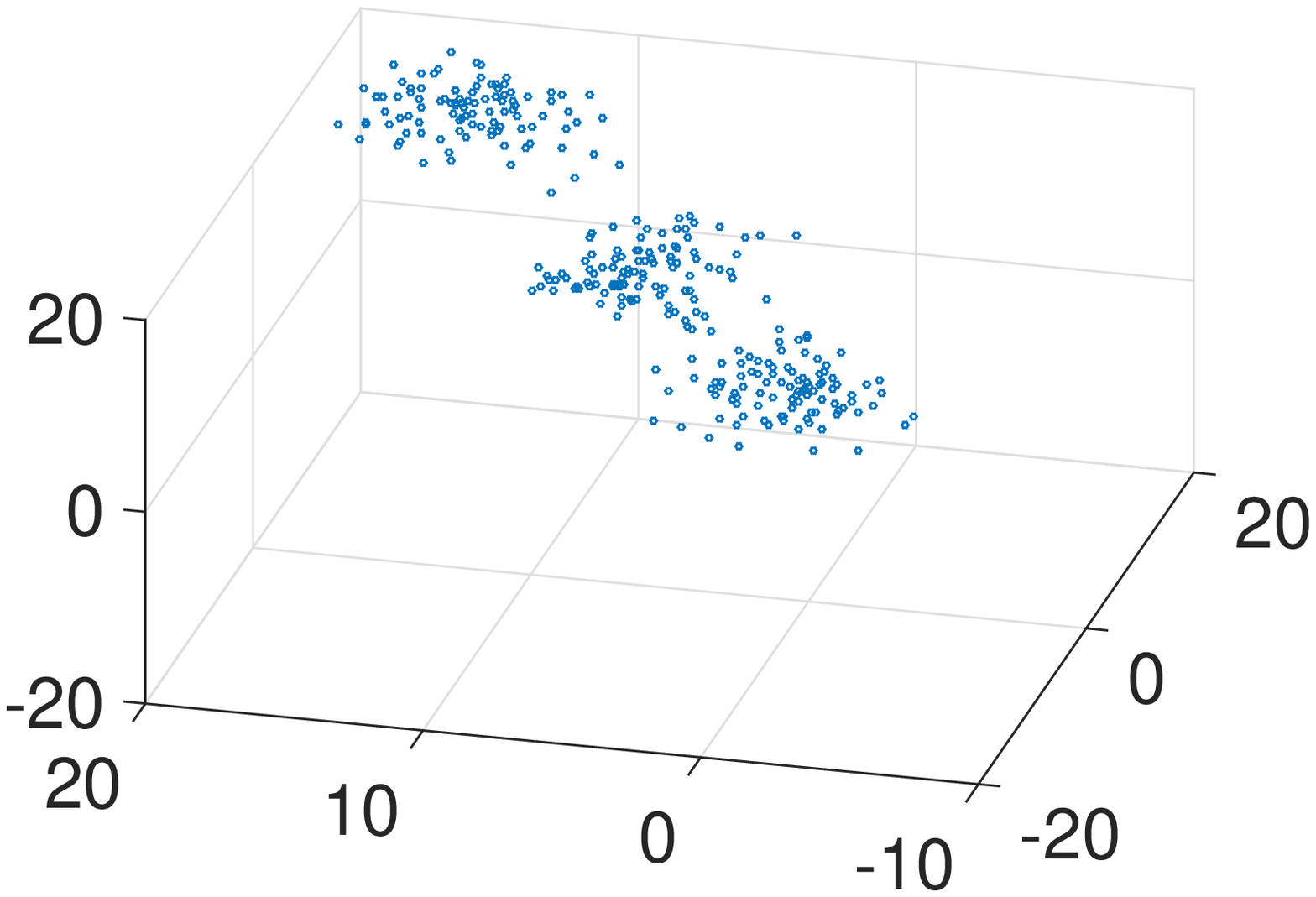}
		\caption{}
		\label{phase1_arg_2}
	\end{subfigure}
	\begin{subfigure}[b]{.24\textwidth}
		\centering
		\includegraphics[width=\textwidth]{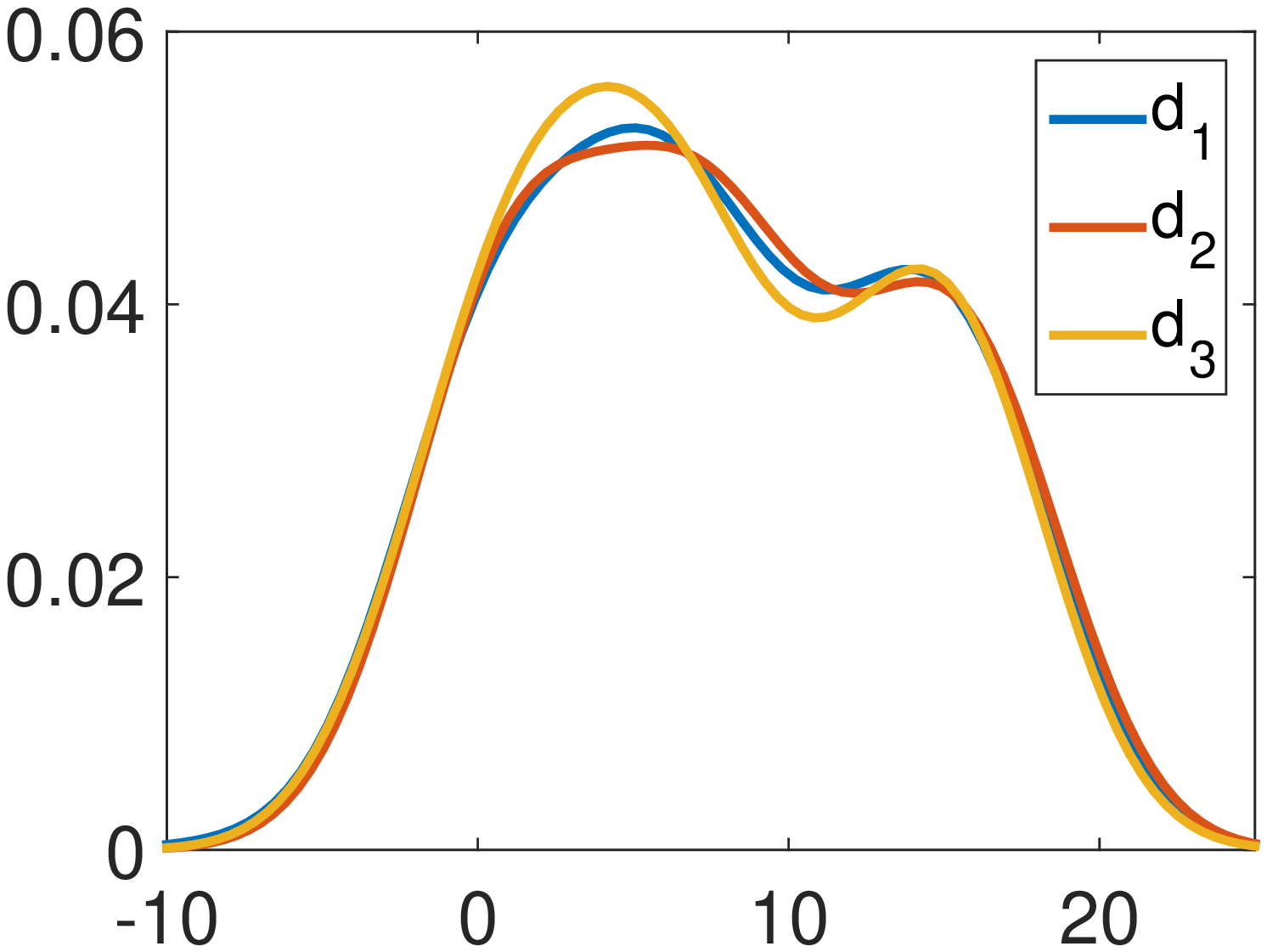}
		\caption{}
		\label{kdensity_2}
	\end{subfigure}
	\vspace{-0.5em}
	\caption{(a) Distribution of 300 points in 3-dimensional space; (b) Estimated normal probability density of 200 points in each dimension; (c) Distribution of 300 points having the same distribution as (a) in each dimension, but with small covariances; (d) Estimated normal probability density of (c).}
	\label{example_1}
	\vspace{-1.5em}
\end{figure*}

Our first phase searches for lower dimensional clusters. These are called \emph{base clusters} as they are the basis that form higher dimensional clusters. Unlike traditional bottom-up subspace clustering algorithms such as CLIQUE \cite{Agrawal:1998:ASC:276304.276314}, ENCLUS and MAFIA~\cite{parsons2004subspace} that search for dense units in individual dimensions, we search for base clusters in subspaces with two or more dimensions. This approach can preserve the covariance between different dimensions. Not only is the proximity between points in each dimension important but also the covariances of values in different dimensions are critical to decide the formation of clusters. Figure \ref{phase1_arg_1} shows a distribution of 300 points in a 3-dimensional space. Points $\{x_i\}_{i=1}^{100}$ are from a normal distribution $\mathcal{N}(1,2)$ and form a dense unit in dimension $d_1$. Similarly, $\{x_i\}_{i=101}^{200}$ and $\{x_i\}_{i=201}^{300}$ follow two normal distributions $\mathcal{N}(7,2)$ and $\mathcal{N}(10,2)$ in $d_2$ and $d_3$, and form two dense units in these dimensions respectively. When clustering these points in $2D$ and $3D$ spaces, where covariance is implicitly implied, these points do not form any cluster, as confirmed by k-means or visual inspection of Figure \ref{phase1_arg_1}. This can be explained with the normal probability density distribution in Figure \ref{kdensity_1}. While the first 100 points $\{x_i\}_{i=1}^{100}$ are close to each other in $d_1$, the same set of points have large variances in $d_2$ and $d_3$, and cannot be considered close in higher dimensional space. The correlation between different dimensions is omitted when each dimension is considered separately.

Figure \ref{phase1_arg_2} shows an example where missing clusters can be prevented. It contains 300 points whose coordinates in each dimension are sampled from three equal size normal distributions $N(1,2)$, $N(7,2)$ and $N(10,2)$. In fact, if we consider each dimension separately, the values in each dimension are the same as the previous distribution shown in Figure \ref{phase1_arg_1}. However, in this example, we enforce that for each point $x_i$, its coordinates in all dimensions must be drawn from the same distribution. No dense units are found in each individual dimension since the points are normally distributed, as can be observed from the probability density distributions in Figure \ref{kdensity_2} (which do not show any significant peaks, compared to Figure \ref{kdensity_1}). With no dense unit, no cluster is found by the aforementioned methods. However, it is visually evident that 3 clusters exist in this dataset.

Note that the dimensionality of the final clusters is higher than $p$ if the search for base clusters starts with a $p$-dimensional subspace. For example, if the algorithm performs phase 1 with 3-dimensional ($3D$) subspaces, it assumes there is no cluster in $2D$ or $1D$ subspaces. For this reason, it is ideal to start phase 1 in subspaces that are low dimensional, i.e., keeping $p$ small. Another factor that affects the algorithm is the number of subspaces that need to be searched. If the dimensionality of the full space is low, it is feasible to perform the search in all of its $p$-dimensional subspaces. As an example with a $50D$ dataset, the total number of $2D$ subspaces is ${{50}\choose{2}} =1225$. If the number of dimensions is high, it is possible to perform sampling of subspaces instead of considering all of them, as long as each dimension is sampled sufficiently frequently. In this paper, we search for base clusters in all $2D$ subspaces if the number of dimensions is less than 100, while in higher dimensional datasets we perform subspace sampling. We find that in practice this provides a good balance between clustering quality and computational complexity.

Table \ref{example_of_z} shows an example of the output of phase 1. Note that we use the following notation: $C_{S_i,j}$ denotes the $j^{th}$ cluster in the subspace $S_i$. It searches for clusters in 6 subspaces $\{S_1,...,S_6\}$ of the full data space $S$. Points $x_1$, $x_2$ and $x_3$ belong to the same cluster $C_{S_1,1}$ in subspace $S_1$. They also belong to cluster $C_{S_2,1}$ in subspace $S_2$, while sharing no common cluster in other subspaces.
\begin{table}[b]
\centering
\begin{tabular}{c|c|c|c|c|c|c|}
\cline{2-7}
                                   & \multicolumn{6}{c|}{Subspaces of base clusters}                                     \\ \hline
\multicolumn{1}{|l|}{Points} & $S_1$ & $S_2$ & $S_3$ & $S_4$ & $S_5$ & $S_6$ \\ \hline
\multicolumn{1}{|c|}{$x_1$}           & $C_{S_1,1}$      & $C_{S_2,1}$      & $C_{S_3,1}$      & $\emptyset$ & 	$\emptyset$ & $C_{S_6,1}$      \\ \hline
\multicolumn{1}{|c|}{$x_2$}           & $C_{S_1,1}$      & $C_{S_2,1}$      & $C_{S_3,2}$      & $C_{S_4,1}$      & 	$\emptyset$ & $C_{S_6,1}$      \\ \hline
\multicolumn{1}{|c|}{$x_3$}           & $C_{S_11}$      & $C_{S_2,1}$      & $C_{S_3,3}$      & 	$\emptyset$ & $C_{S_5,1}$      & $C_{S_6,1}$      \\ \hline
\multicolumn{1}{|c|}{$x_4$}           & $C_{S_1,2}$      & $C_{S_2,2}$      & $C_{S_3,4}$      & $C_{S_4,2}$      & $C_{S_5,1}$      & $C_{S_6,1}$      \\ \hline
\multicolumn{1}{|c|}{$x_5$}           & $C_{S_1,2}$      & $C_{S_2,2}$      & $C_{S_3,4}$      & $C_{S_4,2}$      & $C_{S_5,2}$      & $C_{S_6,1}$      \\ \hline
\end{tabular}
\caption{Base clusters in six subspaces of the dataset.}
\label{example_of_z}
\vspace{-1em}
\end{table}

The base clusters found that cover similar sets of data points are aggregated together to form clusters in higher dimensional subspaces. Subspace $S_i$ of a high dimensional cluster is constituted of all the dimensions of its aggregated base clusters. According to Lemma 1, these base clusters can be considered as evidence to increase the posterior probability of the formation of the high dimensional cluster.

\subsection{Phase 2: High Dimensional Cluster Construction}
Phase 2 learns the patterns of proximity among the points from the output of phase 1, which is denoted as $Z$ (Table \ref{example_of_z}), to derive the final clusters and present them in a succinct and interpretable way. To this end, we consider $Z$ as a transaction database where each point corresponds to a transaction and the base clusters covering that point are the items of that transaction. From Table \ref{example_of_z}, the first row is the transaction of point $x_1$, and the corresponding items are $C_{S_11}, C_{S_21}, C_{S_31}, C_{S_61}$. Subsequently, we use $Z$ as the input to build an FP-Tree \cite{han2000mining}, in which each branch is an aggregation of base clusters and represents a high dimensional cluster. Effectively, each \textit{frequent pattern} mined from the tree indicates a sufficiently large group of points that form clusters in a high dimensional subspace. The minimal size of a cluster is controlled by the minimum support (\verb!min_sup!) \cite{han2000mining} of the frequent pattern mining process. In practice, the choice of the \verb!min_sup! parameter can be guided by the expected minimum cluster size. Note that not all frequent patterns are useful as they can produce redundant clusters. For any cluster defined by the frequent pattern $F_i$, all subsets of $F_i$ are also frequent, and correspond to clusters in lower dimensions, but none of them form a cluster as complete as $F_i$ does. Therefore, we only need to mine the maximal frequent patterns.

In addition, it is important to control the number of frequent patterns since these can quickly grow. Prior to the extraction of maximal frequent patterns, phase 2 analyses the frequencies of patterns at different levels of the FP-Tree, and prunes small branches with low frequencies. These branches correspond to insignificant patterns and only reflect the characteristics of a small portion of the points that do not justify a cluster. This is essential to prevent the algorithm from producing a huge number of small and meaningless clusters. To this end, phase 2 first performs a scan on the FP-Tree and records the frequency on each branch at each depth level of the tree. It then finds the knee-point \cite{zhao2008knee}, which indicates the level after which the frequencies significantly drop. Subsequently, the remainder of that branch is pruned. We present a running example using Table \ref{example_of_z} as the input, with \verb!min_sup! set to 0.4. Figure \ref{fp_tree} shows the FP-Tree before being pruned. The pruning eliminates the node $C_{S_51}$, which has a frequency of 1 (i.e., the patterns only apply to $x_3$) and hence should not justify a separate cluster. The branch that starts at node $C_{S_51}$ on the right branch of the tree is also pruned (the patterns only apply to $x_4$). Eventually, two clusters are found as presented in Table \ref{running_example}.

The process of building the tree and mining maximal frequent patterns only requires two passes over the input $Z$. The process of pruning the tree performs one traversal of the tree, which is linear with respect to the size of $Z$. This contributes to the low computational complexity and therefore improves the scalability of the algorithm.

\begin{table}[]
\begin{minipage}[b]{\linewidth}
\centering
\includegraphics[]{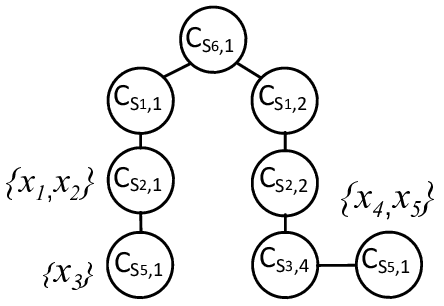}
\captionof{figure}{FP-Tree built from Table \ref{example_of_z}}
\label{fp_tree}
\end{minipage}
\\
\begin{minipage}[b]{\linewidth}
\centering
\begin{tabular}{|l|l|l|}
\hline
& Points             & Patterns                    \\ \hline
$C_1$     & $\{x_1,x_2,x_3\}$ & $\{C_{S_6,1},C_{S_1,1},C_{S_2,1}\}$       \\ \hline
$C_2$     & $\{x_4,x_5\}$      & $\{C_{S_6,1},C_{S_1,2},C_{S_2,2},C_{S_3,4}\}$ \\ \hline
\end{tabular}
\caption{Clusters extracted from FP-Tree of the base clusters shown in Figure \ref{fp_tree}.}
\label{running_example}
\end{minipage}\hfill
\vspace{-2em}
\end{table}

\section{Evaluation}
We evaluate our algorithm using real-life datasets from a variety of applications. First, we apply our algorithm to ten gene expression datasets, and compare its accuracy with six clustering algorithms that are commonly used for biomedical data. Next, we apply the algorithm to a real-life dataset of car parking occupancy in a major city, and quantitatively evaluate the result. Finally, we evaluate the algorithm using synthetic datasets of different sizes and dimensions, and compare the results with traditional bottom-up clustering algorithms \cite{muller2009evaluating} as well as other state-of-the-art subspace clustering algorithms \cite{elhamifar2013sparse,liu2011latent}. We also evaluate the scalability of our algorithm on large datasets. All experiments are conducted with MATLAB on an Intel Core i7-4790 3.6GHz CPU and 16GB of RAM.

\subsection{Clustering Gene Expression Data}
We first perform clustering on ten gene expression datasets that were widely used in different studies \cite{chen2017subspace}. The sizes and characteristics of these datasets are summarised in Table \ref{gene_expression_dataset}. The performance of our proposed algorithm is compared with 7 other algorithms, including EWKM \cite{jing2007entropy}, BBAC-S \cite{banerjee2007generalized}, ITCC \cite{dhillon2003information}, FFCFW \cite{tjhi2006flexible}, HICC \cite{cheng2016hicc}, and SWCC \cite{chen2017subspace}. The metric used to measure the correctness of the result is normalised mutual information (NMI) \cite{larson2010introduction}. Note that we also used precision, recall, f-measure, and accuracy to evaluate the clustering results but do not present the comparison numbers here because they are not directly comparable to those presented in the previous papers~\cite{chen2017subspace}. Our approach to true/false positives and true/false negatives for clustering is slightly different from the one used in the aforementioned papers. After finding the clusters, these algorithms use the Hungarian algorithm \cite{jonker1986improving} to find the best mapping between the clustering result and the given labels. However, the Hungarian algorithm requires that the algorithms find the correct number of clusters, which is guaranteed in \cite{chen2017subspace} because this is given as an input parameter. Our algorithm does not require the number of clusters to be specified in advance, and hence it is not always guaranteed to produce the correct number of clusters. Instead, we use the approach presented in \cite{larson2010introduction} to determine true/false positives and true/false negatives.

\begin{table}[t]
\centering
\begin{tabular}{lllll}
\hline
Abbr. & Name           & \#Patients & \#Genes & \#Classes \\ \hline
ADE   & adenocarcinoma & 76         & 9868    & 2         \\
BRA   & brain          & 42         & 5597    & 5         \\
BR2   & breast.2.class & 78         & 4869    & 2         \\
BR3   & breast.3.class & 96         & 4869    & 3         \\
COL   & colon          & 62         & 2000    & 2         \\
LEU   & leukemia       & 38         & 3051    & 2         \\
LYM   & lymphoma       & 62         & 4026    & 3         \\
NCI   & nci 60         & 61         & 5244    & 8         \\
PRO   & prostate       & 102        & 6033    & 2         \\
SRB   & srbct          & 63         & 2308    & 4         \\ \hline
\end{tabular}
\caption{Characteristics of 10 gene expression datasets.}
\label{gene_expression_dataset}
\vspace{-1em}
\end{table}

Next, we present the parameter settings for the algorithms in this experiment. In phase 1, we start the search for base clusters in two-dimensional subspaces ($2D$), and use k-means to find the base clusters in each of these subspaces. Therefore, there are only two parameters required by our algorithm: the number of base clusters $k$ in each subspace, and the expected minimum size of a cluster, reflected in $min\_sup$. We conducted the experiment with 5 values of $k$ $\{25,20,15,10,5\}$ and 6 values of $min\_sup$ $\{0.25,0.2,0.15,0.1,0.07,0.05\}$, i.e., 30 runs in total. The other algorithms are provided with the correct number of clusters. The full parameter settings of the other methods are described in detail in \cite{chen2017subspace}.

\begin{table}[b]
\vspace{-1em}
\centering
\begin{tabular}{m{0.4cm}|m{0.5cm}|m{0.8cm}m{1cm}m{0.6cm}m{0.6cm}m{0.6cm}m{0.9cm}}
\hline
Data & Ours                                              & EWKM                                                & BBAC-S                                              & ITCC & FFCFW & HICC & SWCC                                                \\ \hline
ADE  & 0.03                                                     & 0.00                                                & 0.00                                                & 0.01 & 0.00  & 0.01 & \cellcolor[HTML]{C0C0C0}0.02                        \\
BRA  & 0.25      & 0.19                                                & \cellcolor[HTML]{333333}{\color[HTML]{FFFFFF} 0.45} & 0.00 & 0.00  & 0.13 & \cellcolor[HTML]{333333}{\color[HTML]{FFFFFF} 0.40} \\
BR2  & 0.05      & 0.01                                                & 0.04                                                & 0.00 & 0.00  & 0.01 & \cellcolor[HTML]{C0C0C0}0.06                        \\
BR3  & 0.09      & \cellcolor[HTML]{9B9B9B}0.08                        & \cellcolor[HTML]{333333}{\color[HTML]{FFFFFF} 0.21} & 0.00 & 0.00  & 0.02 & \cellcolor[HTML]{333333}{\color[HTML]{FFFFFF} 0.22} \\
COL  & 0.07                                                     & 0.02                                                & 0.04                                                & 0.00 & 0.00  & 0.01 & 0.04                                                \\
LEU  & 0.13      & \cellcolor[HTML]{333333}{\color[HTML]{FFFFFF} 0.21} & \cellcolor[HTML]{333333}{\color[HTML]{FFFFFF} 0.43} & 0.00 & 0.00  & 0.05 & \cellcolor[HTML]{333333}{\color[HTML]{FFFFFF} 0.34} \\
LYM  & 0.2       & \cellcolor[HTML]{333333}{\color[HTML]{FFFFFF} 0.33} & \cellcolor[HTML]{333333}{\color[HTML]{FFFFFF} 0.62} & 0.00 & 0.07  & 0.04 & \cellcolor[HTML]{333333}{\color[HTML]{FFFFFF} 0.47} \\
NCI  & 0.3       & 0.23                                                & \cellcolor[HTML]{333333}{\color[HTML]{FFFFFF} 0.61} & 0.00 & 0.24  & 0.15 & \cellcolor[HTML]{333333}{\color[HTML]{FFFFFF} 0.53} \\
PRO  & 0.05                                                    & 0.01                                                & 0.02                                                & 0.00 & 0.00  & 0.01 & 0.03                                                \\
SRB  & 0.25      & 0.14                                                & \cellcolor[HTML]{C0C0C0}0.26                        & 0.00 & 0.00  & 0.07 & 0.20                                                \\ \hline
\end{tabular}
\caption{Comparison of clustering results (using NMI) of our algorithm with 6 other clustering algorithms. A white cell indicates a worse result than our algorithm, a black cell indicates a better result, a grey cell shows no statistical differences between results.}
\label{clustering_result}
\vspace{-2em}
\end{table}

We compute NMI for each clustering result and compare the average results of all algorithms in Table \ref{clustering_result}. A t-test \cite{kim2015t} is performed with a significance level of 5\% to determine if the average NMI values produced by our algorithm are significantly different from those produced by the other algorithms. In Table \ref{clustering_result}, the cells of the other algorithms are color-coded to highlight the relative performance of our algorithm. A white cell of a baseline algorithm indicates that the baseline algorithm performs worse than ours with statistical significance, a black cell indicates the baseline algorithm has a higher NMI value than ours, whereas a grey cell shows no statistical difference between the results. For example, the last row of the table indicates that the result of our algorithm is better than most of the other algorithms, has no statistical difference compared to BBAC-S, and is worse than k-means. It can be observed from the results that our algorithm produces comparable or better results than all other algorithms for the datasets of ADE, BR2, COL, PRO, and SRB (except for k-means). Our algorithm also performs better than ITCC, FFCFW, and HICC on all datasets.

In summary, this demonstrates that we can achieve as good or better accuracy than state-of-the-art algorithms over a variety of genomic datasets.

\subsection{Clustering Car Parking Occupancy Data}
Next, we demonstrate the capability of our algorithm to work with data collected from a real-life IoT application.

The City of Melbourne has deployed sensors to record parking events at parking bays around the central business district (CBD). We extract the start and end time of all parking events to compile the parking occupancy at 276 locations at 15 minutes intervals between 09:00-18:00, yielding an input of size $276 \times 36$ for each day. The aim is to find clusters of car parking spots that have similar patterns of occupancy at certain times of the day. Each clustering task is performed on five days worth of data to find the patterns of parking occupancy during weekdays. Parking occupancy is an important metric that indicates the efficiency of car park utilisation \cite{kelly2006influence}, which heavily affects traffic, ease of commute and business in the CBD. Analysing the car occupancy can reveal patterns in parking behaviour at different car parks during different times of the day, which can then be used to review the parking hotspots or tariffs.

\begin{figure}[t]
	\centering
	\begin{subfigure}[b]{.49\textwidth}
		\centering
		\includegraphics[width=\textwidth]{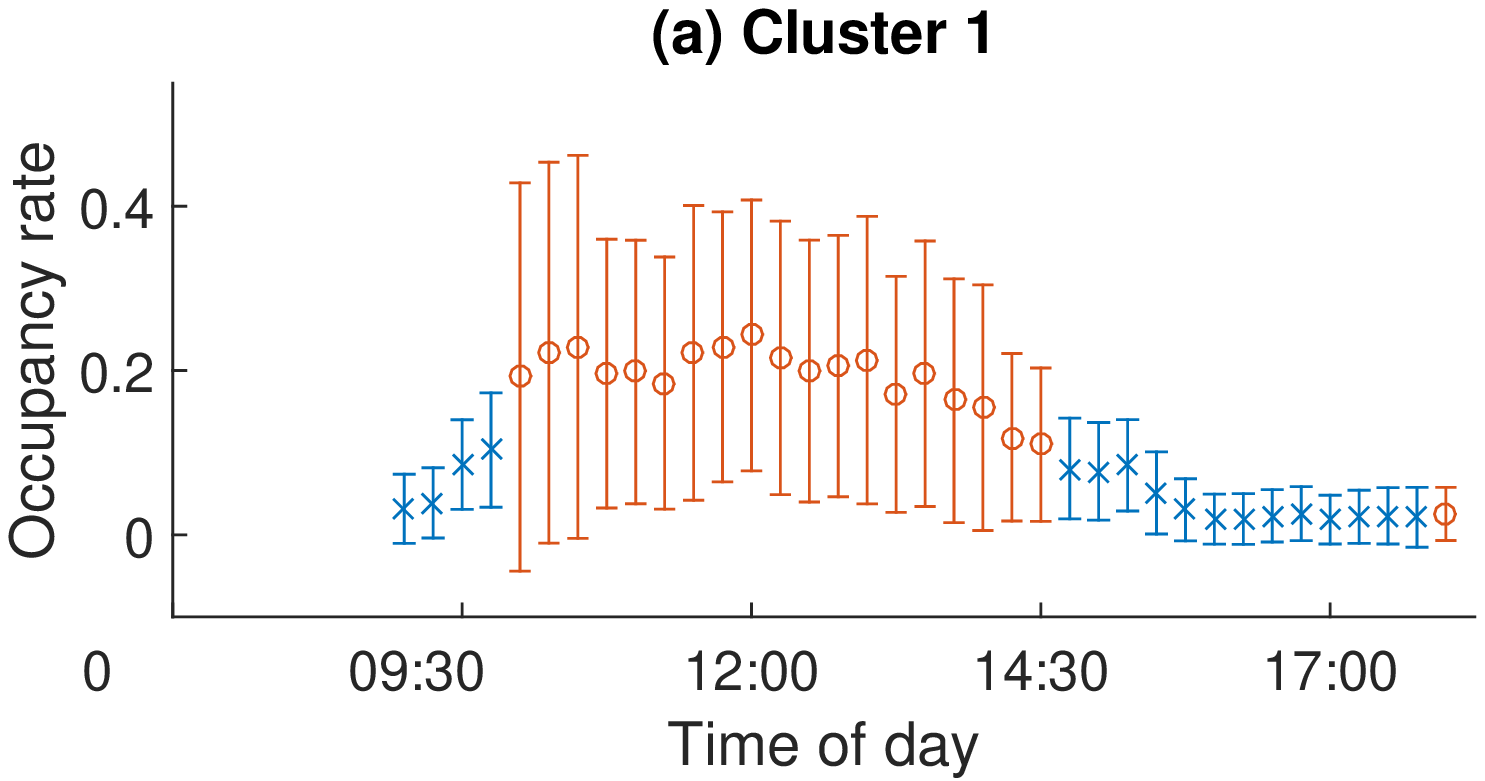}
	\end{subfigure}
	\begin{subfigure}[b]{.49\textwidth}
		\centering
		\includegraphics[width=\textwidth]{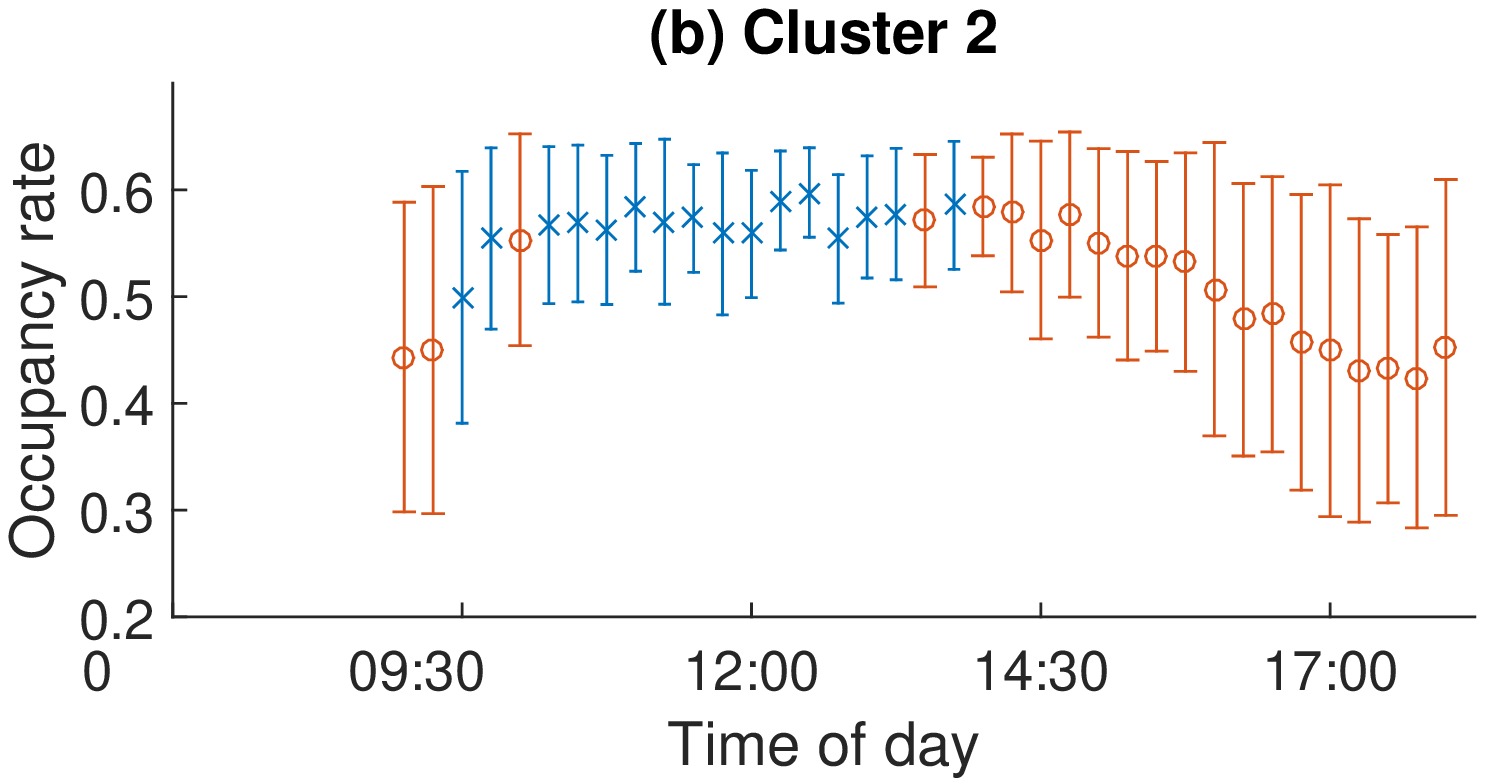}
	\end{subfigure}
	\caption{Verification of cluster quality.}
	\label{parking_cluster_quality}
	\vspace{-1.5em}
\end{figure}

By clustering the parking occupancy data, each cluster $C_{S_i}^{X_j}$ represents a parking pattern observed at the locations (points) $X_j$ during the times (dimensions) defined by $S_i$. The results are evaluated using two methods. First, we analysed the coherence of each cluster by statistically verifying whether the clustered parking bays have small deviations in the values of parking occupancy during the corresponding time periods, compared to the rest of the data. The examples of two clusters are shown in Figure \ref{parking_cluster_quality}, where each blue bar represents the mean and standard deviation of the parking occupancy at a certain time of the day, observed at parking bays grouped by the cluster. For example, \verb!Cluster 1! in Figure \ref{parking_cluster_quality}a shows the pattern shared by a group of parking bays during 9:00-10:30 and 14:45-17:45 with small standard deviations, compared to significant deviations at other times of the day. Similarly, \verb!Cluster 2! shows another pattern that has an occupancy rate of 55\% around midday, while such correlation is not observed at other times of the day.

Second, to quantify the effectiveness of the method, we use the clusering result to construct an ensemble prediction model to predict the parking occupancy over the next few hours, and compare the accuracy of our model with other models. The details of the prediction models are as follows:

\begin{itemize}
	\item Model 1 applies decision tree regression \cite{xu2005decision} directly on the occupancy data.
	\item Model 2 first clusters the data using the proposed algorithm and then fits a decision tree regression on the set of car parks in each cluster separately.
	\item Model 3 follows the same approach as Model 2 except that it uses the k-means algorithm in the first phase.
\end{itemize}

Each cluster ideally represents a pattern of parking occupancy shared by a group of parking bays. Fitting a submodel to each cluster allows each submodel to learn the data in more detail and predict with higher accuracy if the values are coherent. Therefore, the accuracy of the prediction model directly reflects the quality of the clusters. This approach of using clustering in an ensemble prediction model has previously been used in \cite{BENMOUIZA2013561,fahiman2017improving}.

\begin{figure}[t]
	\centering
	\begin{subfigure}[b]{.49\textwidth}
		\centering
		\includegraphics[width=\textwidth,height=4.5cm]{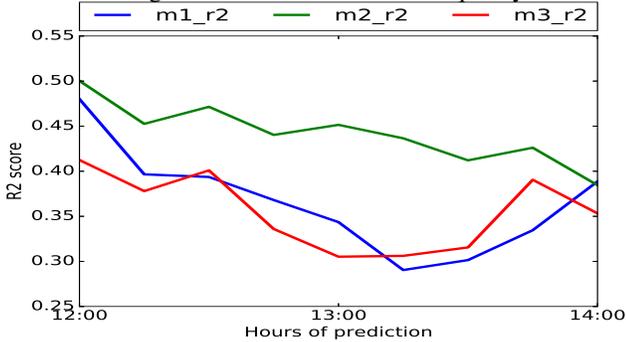}
	\end{subfigure}
	\caption{Prediction accuracy of 3 models.}
	\label{prediction_accuracy}
	\vspace{-1em}
\end{figure}

Each prediction model uses the values between 09:00-12:45 as training data to predict the occupancy rates of the next two hours. The coefficient of determination (R2) \cite{ozer1985correlation} is used to measure the accuracy. Figure \ref{prediction_accuracy} shows that our model ($m2$) outperforms the other two, reflected in higher R2 scores. It can also be observed that Model 3, which relies on k-means, is not as accurate as Model 1, which implies that fitting submodels to the input does not always translate to higher accuracy. In fact, the accuracy can deteriorate if the values in each submodel are not coherent.

In summary, by incorporating the clustering results into decision tree regression to improve the prediction accuracy, we quantitatively show that our clustering algorithm can cluster data into meaningful partitions that share similar patterns. It also demonstrates its capability of handling real datasets with high levels of noise and outliers.

\subsection{Experiments with Synthetic Data}

\begin{figure}[!ht]
	\captionsetup[subfigure]{}
	\begin{subfigure}[b]{.23\textwidth}
		\centering
		\includegraphics[width=\textwidth,height=3.7cm]{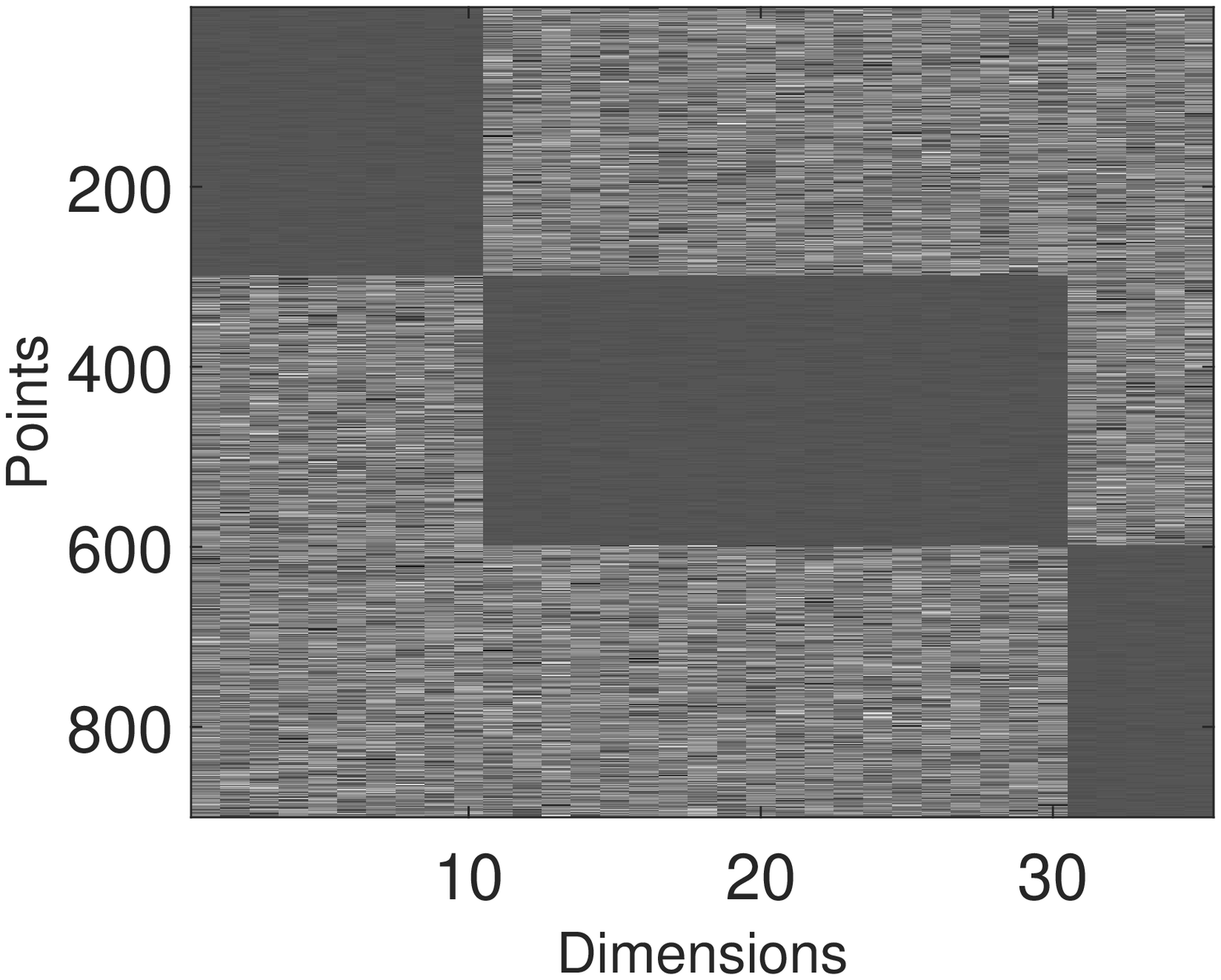}
		\caption{}
		\label{disjoint_subspace_1}
	\end{subfigure}
	\begin{subfigure}[b]{.23\textwidth}
		\centering
		\includegraphics[width=\textwidth,height=3.7cm]{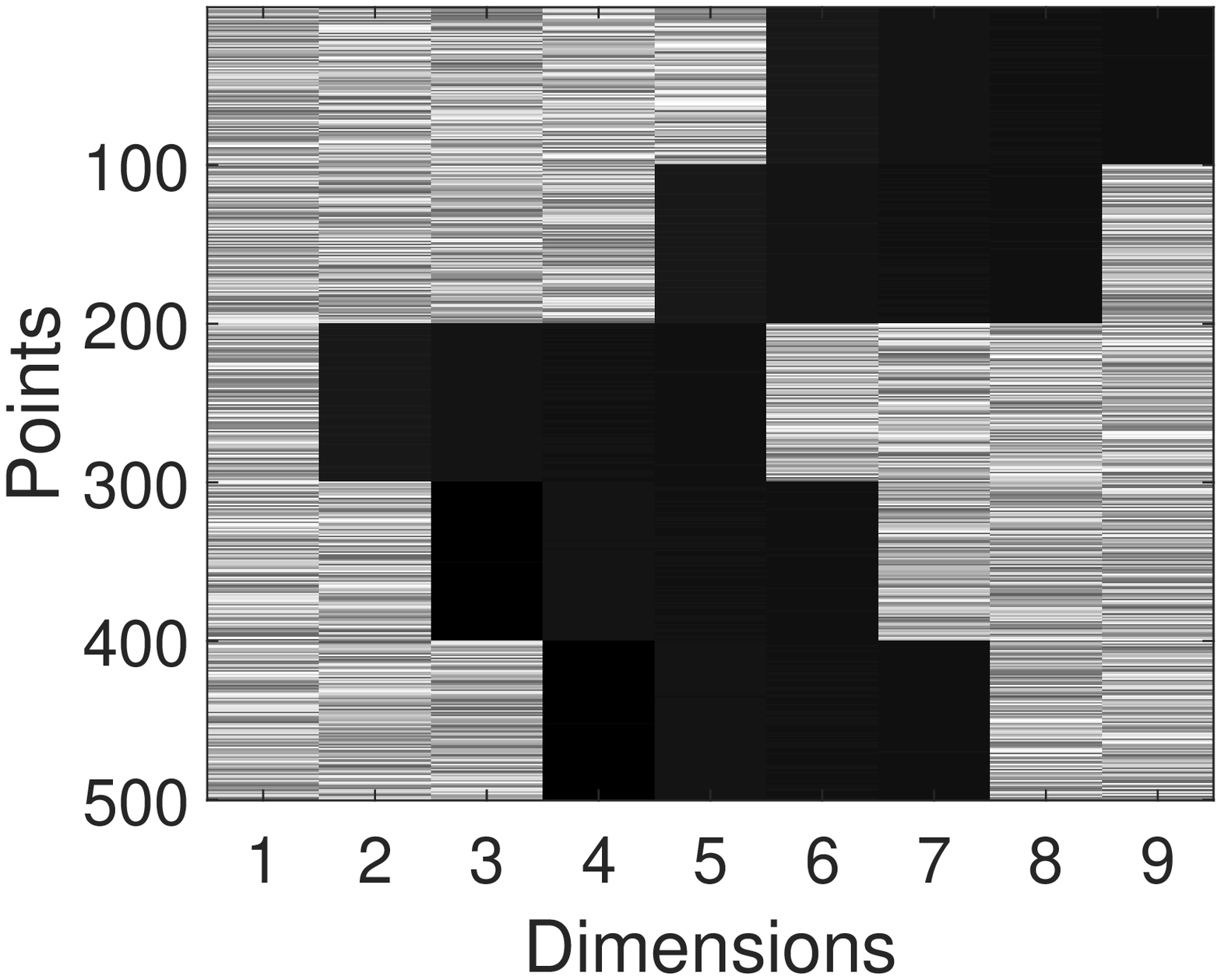}
		\caption{}
		\label{non_disjoint_1}
	\end{subfigure}

	\caption{(a) Clusters in disjoint subspaces with no outliers. Points $\{x_i\}_{i=1}^{300}$, $\{x_i\}_{i=301}^{600}$, $\{x_i\}_{i=601}^{900}$ form 3 clusters in 3 different subspaces, (b) Clusters in non-disjoint subspaces.}
	\label{synthetic_data}
	\vspace{-1em}
\end{figure}

We further evaluate our algorithm on a variety of synthetic datasets in order to assess (1) its capability to find clusters in disjoint and non-disjoint subspaces, and (2) its capability to scale with large inputs. Figure \ref{synthetic_data} shows the grayscale heatmap of a sample dataset containing 900 points in a 35-dimensional space. The points $\{x_i\}_{i=1}^{300}$ form a cluster in the subspace $S_{1:10}$, which is constituted by the dimensions $\{d_1,...d_{10}\}$; points $\{x_i\}_{i=301}^{600}$ form a cluster in the subspace $S_{11:30}$, which is constituted by the dimensions $\{d_{11},...,d_{30}\}$. The points within the same clusters are more coherent, which is reflected by the more uniform shade of gray of the heatmap. These two clusters intersect only at the origin and hence are disjoint. On the other hand, Figure \ref{non_disjoint_1} is an example of data having clusters residing in non-disjoint subspaces, in which \verb!Cluster 1! spans subspace $S_{6789}$ and \verb!Cluster 2! spans subspace $S_{5678}$.

In this experiment, we start the search for base clusters in $2D$ subspaces. k-means is used to find base clusters in phase 1. Two parameters are required for our algorithm, which is the number of base clusters $k$ in each subspace, and the minimum support \verb!min_sup! required for the construction of the FP-Tree. Note that the value of \verb!min_sup! can be deduced from the minimum expected number of points of a cluster. Setting an appropriate value for $k$ is non-trivial. As we argued earlier, the purpose of phase 1 is to find the similarity in cluster membership of the points in the low dimensional subspaces, rather than the exact cluster of each point. We invoke 12 iterations of our algorithm with $k\in\{3,5,10,15,20,25,30,35,40,45,50, 55\}$ and take the best result. For the baseline algorithms, we also analyse the properties of the synthetic data to derive the data density, the correct number of clusters, and the average dimensions of clusters to provide the ideal range of parameters. The parameters for CLIQUE, SUBCLU, DOC, and STATPC are replicated from \cite{muller2009evaluating}. Each of the baseline algorithms is executed 30 times and the average results are recorded.

\subsubsection{Initial Tests against Baseline Algorithms}
In this section we benchmark our algorithm with clustering algorithms including CLIQUE, SUBCLU, DOC, P3C, and STATPC \cite{muller2009evaluating}, as well as state-of-the-art algorithms including SSC \cite{elhamifar2013sparse}, LRR \cite{liu2011latent}, and SSWC \cite{chen2017subspace}. The number of points of the datasets is set to 1000 and the number of dimensions varies from 10 to 100. The running time limit of each algorithm is set to 30 minutes. The result is summarized in Table \ref{synthetic_data_result}. It can be observed that our algorithm produces comparable or better results compared to SSC and SWCC across all the datasets. These three algorithms, along with STATPC, are the only algorithms that can run to completion within the time threshold. DOC gives consistently high accuracy provided that all five parameters of the algorithm are well-tuned. However, it has significantly higher running time and cannot cluster data larger than $1000 \times 40$ within 30 mintues.

\begin{figure*}[ht]
	\centering
	\begin{subfigure}[t]{.32\textwidth}
		\centering
		\includegraphics[width=\textwidth,height=3.5cm,valign=t]{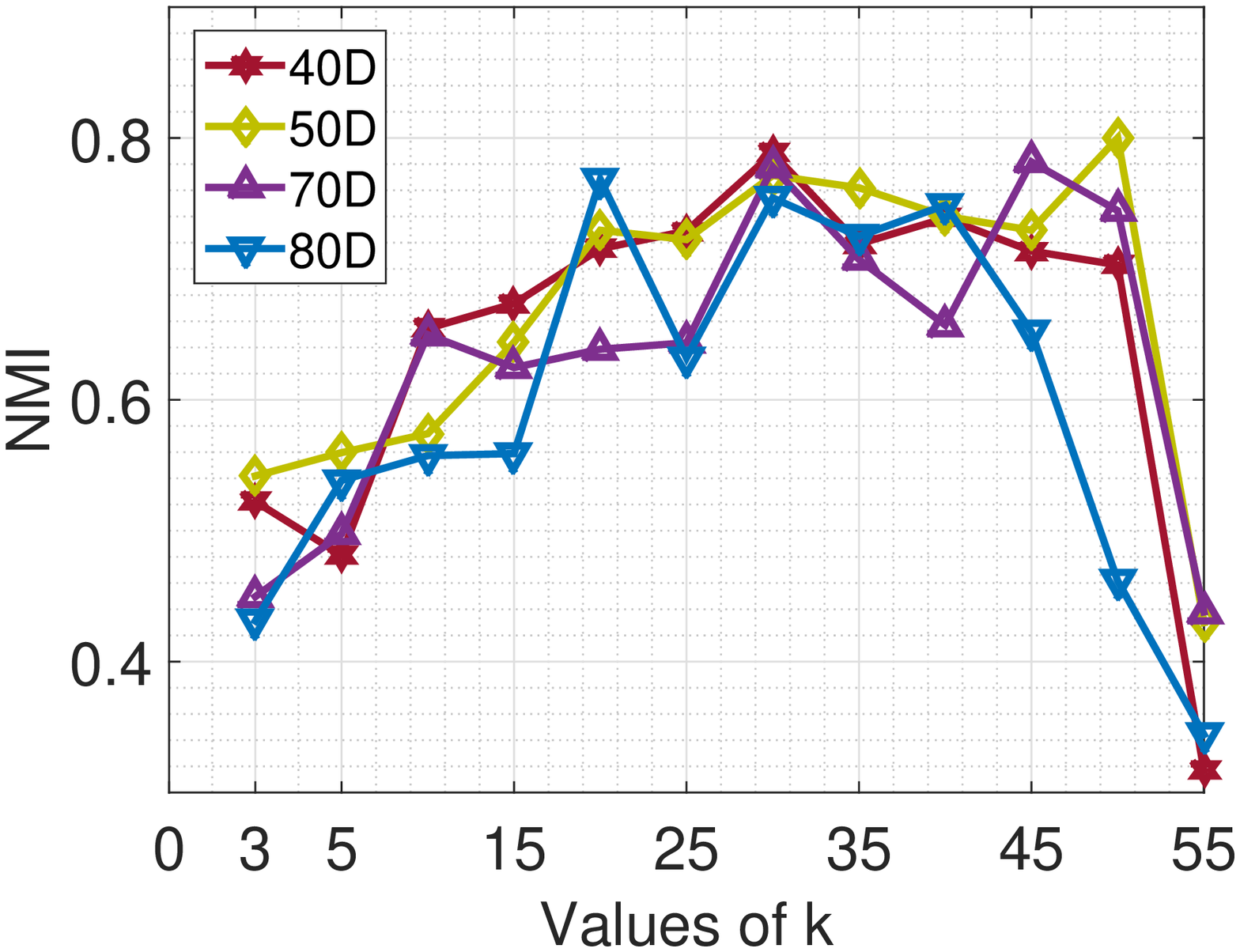}
		\caption{Evaluation of the algorithm sensitivity to the parameter $k$ on synthetic datasets.}
		\label{sensitivity_of_k}
	\end{subfigure}
	\begin{subfigure}[t]{.32\textwidth}
		\centering
		\includegraphics[width=\textwidth,height=3.3cm,valign=t]{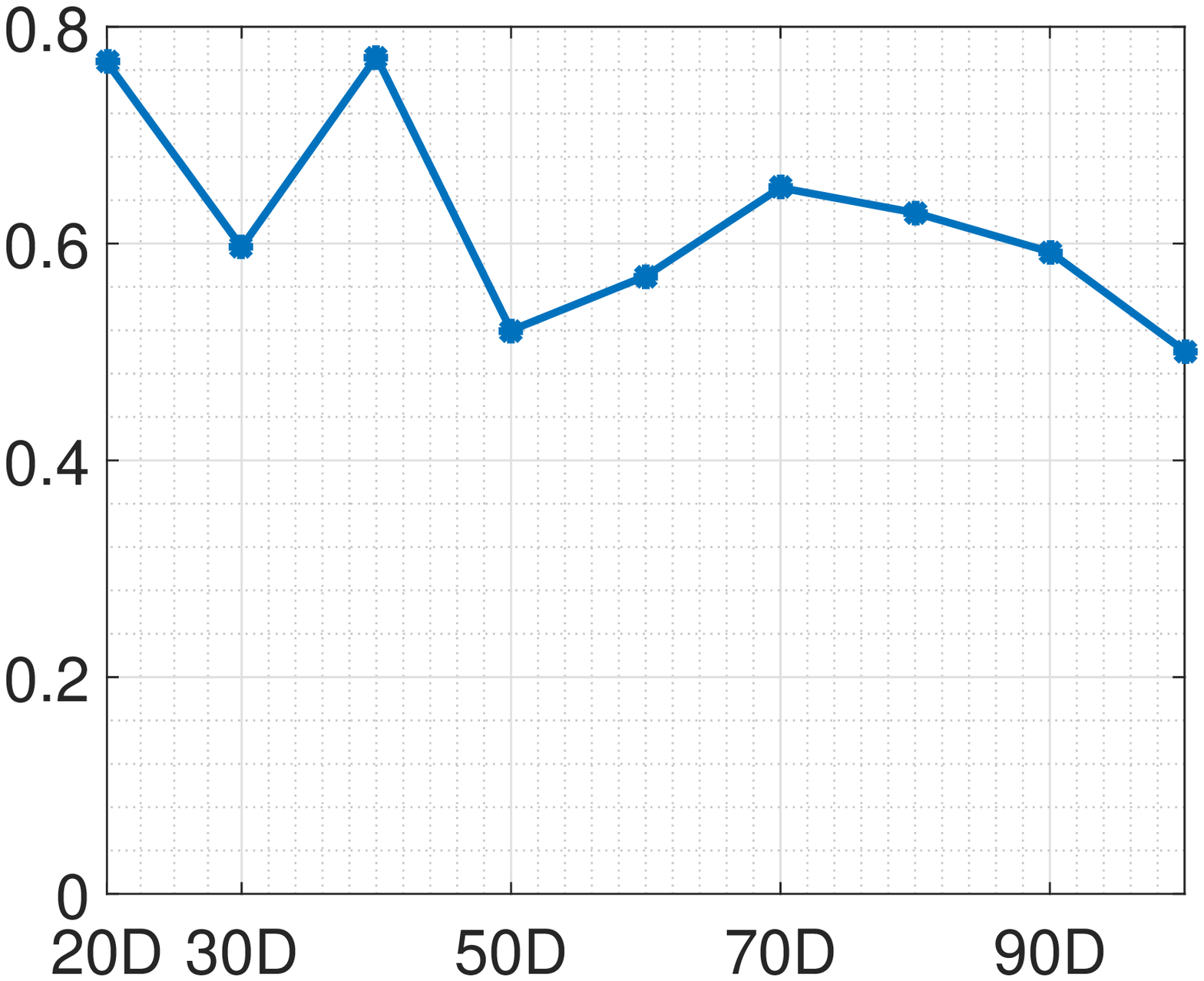}
		\vspace{0.45em}
		\caption{NMI values of finding clusters in non-disjoint subspaces.}
		\label{clustering_non_disjoint}
	\end{subfigure}
	\begin{subfigure}[t]{.32\textwidth}
		\includegraphics[width=\textwidth,height=3.5cm,valign=t]{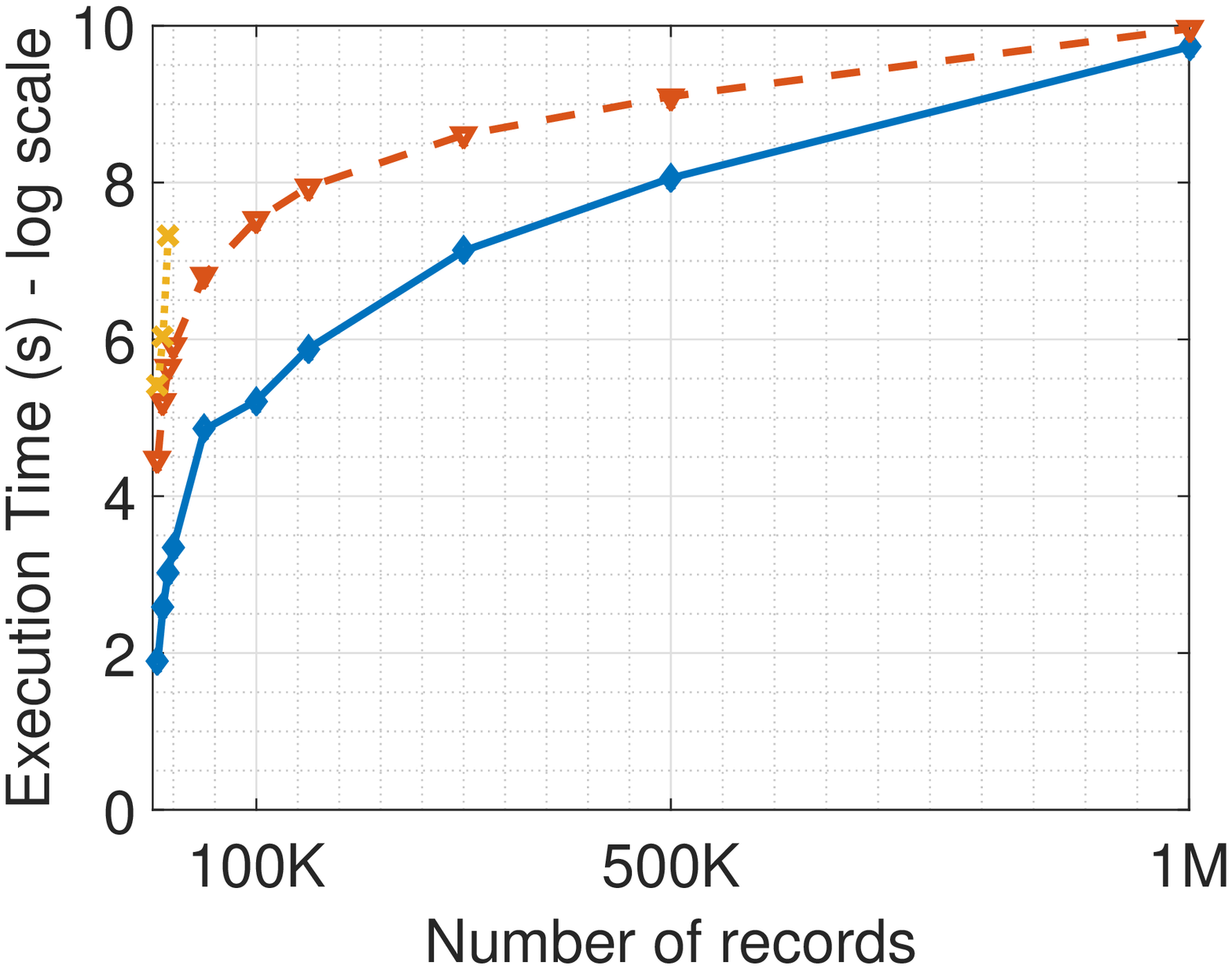}
		\caption{Scalability with number of records}
		\label{execution_time}
	\end{subfigure}
	\caption{Evaluation with synthetic data.}
	\label{evaluation_synthetic_data}
	\vspace{-2em}
\end{figure*}

We also analyse the effect of the setting for the parameter $k$ on the clustering results, as shows in Figure \ref{sensitivity_of_k}. This shows that the clustering results of our algorithm are reasonably insensitive to the setting of $k$ over a wide range of values. For each dataset, there is a value of $k$ at which the clustering result peaks, after which the result deteriorates. We can also observe there is a wide range of $k$ values for which the clustering results are reasonably stable. In practice, the algorithm can be set to run multiple times with different parameters to find the ideal setting.

\begin{table}[H]
\centering
\bgroup
\def\arraystretch{1}
\begin{tabular}{p{1cm}p{0.3cm}p{0.3cm}p{0.3cm}p{0.3cm}p{0.3cm}p{0.3cm}p{0.3cm}p{0.3cm}p{0.3cm}p{0.3cm}}
\cline{1-11}
Algo     & 10D           & 20D          & 30D           & 40D           & 50D           & 60D           & 70D           & 80D           & 90D           & 100D          \\ \hline
CLIQUE   & 0.72          & 0.47         & 0.41          & 0.25          & 0.29          & 0.21          & 0.31          & 0.10          & 0.15          & 0.19          \\
SUBCLU   & 0.64          & -            & -             & -             & -             & -             & -             & -             & -             & -             \\
DOC      & 0.82          & 0.85         & \textbf{0.87} & -             & -             & -             & -             & -             & -             & -             \\
STATPC   & 0.8           & 0.7          & 0.71          & 0.68          & 0.73          & 0.71          & 0.77          & 0.73          & 0.67          & 0.7           \\
SSC      & 0.82          & 0.79         & 0.73          & 0.67          & 0.73          & \textbf{0.87} & \textbf{0.79} & 0.73          & \textbf{0.83} & 0.73          \\
LRR      & 0.79          & 0.87         & 0.74          & 0.6           & 0.25          & 0.16          & 0.62          & -          & -          & -          \\
SWCC     & 0.79          & \textbf{0.9} & 0.84          & 0.75          & 0.73          & 0.74          & 0.61          & 0.76          & 0.52          & \textbf{0.81} \\
Our algo & \textbf{0.86} & 0.88         & 0.84          & \textbf{0.78} & \textbf{0.81} & 0.80           & \textbf{0.79} & \textbf{0.77} & 0.79          & 0.75          \\ \hline
\end{tabular}
\egroup
\caption{Evaluation of algorithms on synthetic datasets (using NMI). The best result for each dataset is highlighed.}
\label{synthetic_data_result}
\vspace{-1em}
\end{table}

\subsubsection{Clustering Non-disjoint Subspaces}
We verify the capability of our algorithm to find clusters in non-disjoint subspaces. In this evaluation we use 1000 data points, where the number of dimensions varies between 20 and 100, and the clusters reside in overlapping subspaces, as illustrated in Figure \ref{non_disjoint_1}. The other algorithms that produce comparable results in the previous section are not included since they are not able to find clusters in overlapping subspaces: SSC and LRR are only able to find clusters in disjoint subspaces \cite{elhamifar2013sparse}\cite{liu2011latent}. Moreover, SWCC assigns weights for each column according to its membership to all clusters and the weights of each column are summed up to 1. This indicates that the memberships of each column to different clusters are exclusive. The result of this evaluation is presented in Figure \ref{clustering_non_disjoint}. The consistently high NMI values ($\ge 0.5$) confirm the capability of the proposed algorithm in finding clusters in non-disjoint subspaces.

\subsubsection{Scalability Tests against SSC and SWCC}
We evaluate the scalability of our algorithm to the number of data points by generating data having 10 dimensions and varying the number of data points from 1,000 to 1,000,000. We include only SSC and SSWC in this scalability evaluation because they are the fastest baseline algorithms with high accuracy. The execution time is presented in Figure \ref{execution_time}. It shows that our algorithm and SWCC can cluster up to 1 million data points while SSC triggers memory errors when the number of points exceeds 15,000.

In summary, these tests on the synthetic datasets demonstrate that our algorithm is relatively insensitive to the choice of parameter settings, while achieving the best overall performance as the number of data points increases.

\section{Conclusion}
We proposed a subspace clustering algorithm to find clusters in non-disjoint subspaces. Unlike traditional bottom-up clustering algorithms, our algorithm starts the search for base clusters in low dimensional subspaces instead of in individual dimensions, in order to capture the covariances of values between dimensions, and to increase the tolerance of the algorithm to variations in the parameter settings.
Our algorithm aggregates the base clusters to form clusters in higher dimensional subspaces based on the technique of frequent pattern mining. Our approach not only avoids the combinatorial complexity of existing bottom-up algorithms, but also ensures more meaningful clustering results by keeping the numbers of final clusters tractable. Our experiments show that the proposed algorithm finds subspace clusters with high accuracy and scales to large inputs, in terms of both the number of records and the number of dimensions. This makes the algorithm practical to many applications in real life, as demonstrated in clustering gene expression data and car parking occupancy data.

\bibliographystyle{IEEEtran}
\bibliography{cikm}

\end{document}